\documentclass[10pt,twocolumn,letterpaper]{article}

\usepackage{cvpr}
\usepackage{times}
\usepackage{epsfig}
\usepackage{graphicx}
\usepackage{amsmath}
\usepackage{amssymb}
\usepackage{cite}
\usepackage{amsmath,amssymb,amsfonts}
\usepackage{algorithmic}
\usepackage{graphicx}
\usepackage{textcomp}
\usepackage{xcolor}
\usepackage{soul}

\def\BibTeX{{\rm B\kern-.05em{\sc i\kern-.025em b}\kern-.08em
		T\kern-.1667em\lower.7ex\hbox{E}\kern-.125emX}}

\usepackage[breaklinks=true,bookmarks=false]{hyperref}

\cvprfinalcopy 

\def\cvprPaperID{****} 

\ifcvprfinal\fi
\begin{document}

\title{ Enabling Incremental Knowledge Transfer for Object Detection at the Edge}

\author{Mohammad Farhadi\\
Arizona State university\\
{\tt\small mfarhadi@asu.edu}
\and
Mehdi Ghasemi\\
Arizona State university\\
{\tt\small mghasem1@asu.edu}
\and
Sarma Vrudhula\\
Arizona State university\\
{\tt\small svrudhul@asu.edu}
\and
Yezhou Yang\\
Arizona State university\\
{\tt\small yz.yang@asu.edu}
}

\maketitle
\thispagestyle{empty}
\begin{abstract}
Object detection using deep neural networks (DNNs) involves a huge amount of computation which impedes its implementation on resource/energy-limited user-end devices. The reason for the success of DNNs is due to having knowledge over all different domains of observed environments. However, we need a limited knowledge of the observed environment at inference time which can be learned using a shallow neural network (\textit{SHNN}). 
In this paper, a system-level design is proposed to improve the energy consumption of object detection on the user-end device. An SHNN is deployed on the user-end device to detect objects in the observing environment. Also, a knowledge transfer mechanism is implemented to update the SHNN model using the DNN knowledge when there is a change in the object domain. DNN knowledge can be obtained from a powerful edge device connected to the user-end device through LAN or Wi-Fi. Experiments demonstrate that the energy consumption of the user-end device and the inference time can be improved by 78\% and 71\% compared with running the deep model on the user-end device. 
\end{abstract}
\section{Introduction}
\label{intro}

The Internet of Things (IoT) refers to a world in which almost any \textit{thing} is instrumented with sensors, computers and communication devices. These \textit{embedded} systems will be utilized in a wide range of domains including  surveillance, retail, healthcare, transportation, industrial robotics and many more. The emergence of IoT is taking place alongside a radical change in how the captured data is processed, namely, with the use of deep neural networks (DNNs). They have become the dominant algorithmic framework for extracting valuable information from massive amounts of disparate data for the purpose of prediction, classification and decision making. 

DNNs are computationally intensive algorithms that involve several layers of nodes that perform billions of multiply-accumulate operations on very large dimensional data sets. Thus, DNNs have to be executed on high performance, large capacity cloud servers. Unfortunately, this means that the massive amounts of data generated by the IoT devices need to be transferred to the cloud, which will soon make this approach infeasible due to the limited bandwidth, the unacceptably large latency, and the potential for compromising the security.  

The preferred solution is to have some or all the data processed by a \textit{user-end} device, which is the first recipient of the data (e.g. a smartphone \cite{mao2019mobieye}, or a smart surveillance camera \cite{xiong2019vikingdet}). However, the limited computation and storage capabilities and/or energy capacity of user-end devices precludes them from executing complex DNN algorithms \cite{farhadi2019novel,han2016mcdnn}.  Edge computing is aimed at addressing this problem by having a part or all of the computation performed on a more powerful local computer called an \textit{edge} device that is connected to the user-end device via a local area network (LAN) or Wi-Fi connection.

\begin{figure}[]
    \centering
    \includegraphics[trim=4.9cm 5.0cm 4.9cm 1.7cm, clip=true,width=8.3cm]{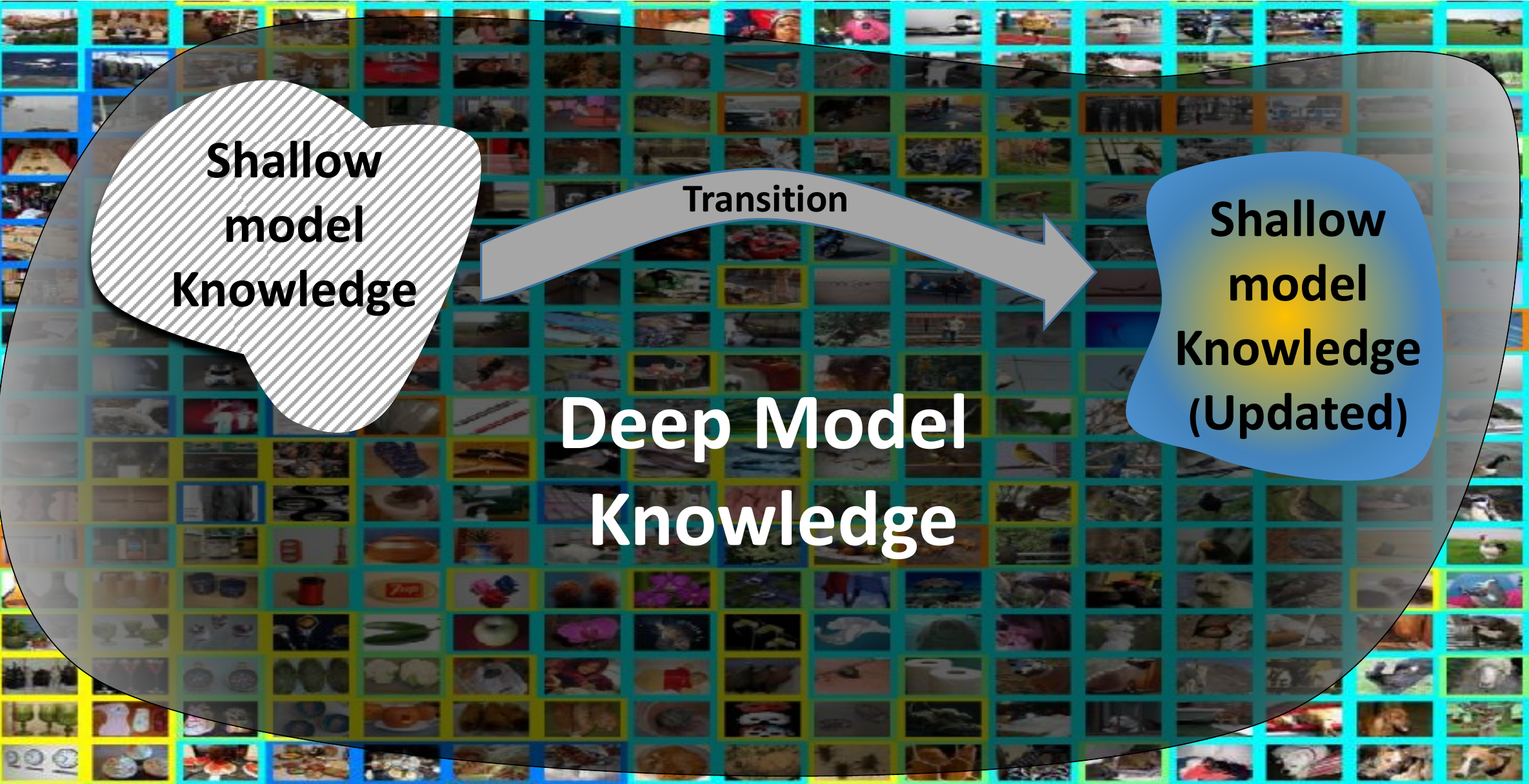}
    \caption{The limited knowledge of shallow model can be adapted to the new environment using the deep model knowledge.}
    \label{motivation}
    
\end{figure}

In this paper, we present a new approach to improve the execution of DNN algorithms in an edge computing environment.  The proposed system targets complex DNN algorithms (e.g., RetinaNet~\cite{lin2017focal}, Faster-RCNN~\cite{ren2015faster}) designed for \textit{object detection} in digital images and videos in different domains (Figure \ref{intro}). Object detection arises in practically every computer vision task and now plays a central role in nearly every one of the applications domains mentioned above. The goal of this work is to demonstrate how two devices~--~a lower performance user-end device and a much higher performance edge device  can cooperate in the execution of computationally intensive DNN algorithms to achieve substantial improvement in energy consumption of the user-end device while achieving nearly the same quality of results as would be if the algorithms were executed solely on the more powerful device.  

The proposed approach is based on a two-level hierarchy of models~--~a \textit{Shallow} neural network (SHNN) (the \textit{student}) that runs on the user-end device, and a DNN (the \textit{oracle}) that runs on the more powerful edge device.  The edge device can also execute the SHNN. The use of the SHNN exploits an important characteristic of images, namely, that in any given image over a period of time, the diversity of objects is quite limited~\cite{farhadi2019tkd}, and therefore, a DNN may not be necessary and a smaller, shallower model will suffice. On the other hand, when changes do occur, they must be detected, and the shallow model must be updated. 

As shown in Figure \ref{motivation}, our approach detects such changes, activates the DNN as required, which in turn transfers the new knowledge (the encoding of the ground truth in the new weights) from the edge device to the user-end device to update the shallow model.  All of this is done while performing inference, i.e., at run-time. The knowledge transferred includes the weights in the decoder layers (layers that detect the objects using the extracted features from previous layers). This transfer enables us to improve the inference time and energy consumption while having a tolerable accuracy loss compared to the deep model.

We demonstrate these ideas by implementing the proposed approach on a pair of devices where the user-end device is an NVIDIA Jetson Nano development kit and the edge device is a Dell workstation with NVIDIA Titan Xp GPU. The experiments show that the proposed method can achieve the desired accuracy with significantly lower inference time. Moreover, the total energy consumption of the user-end device was reduced by \textbf{78\%} when compared to running the DNN entirely on the user-end device. Moreover, the results show that the ratio of object detection accuracy to the energy consumption is improved significantly using the proposed approach. 

\subsection{Contributions}
The main contributions of this paper can be summarized as follows:
\begin{itemize}
    \item We demonstrate a novel framework for transferring knowledge between two devices, one executing a DNN, and the other executing an SHNN. This approach substantially advances edge computing for performing very complex and computationally intensive applications. 
    \item We present an extensive exploration of various ways in which the knowledge transfer can take place, and evaluate them in terms of specific, well-defined metrics. 
    \item The quality of the detection results depends on when the DNN should be activated for possible knowledge transfer.  Toward this, we present a novel \textit{key frame} selection mechanism that significantly improves the efficiency of the knowledge transfer. 
    \item The proposed framework for incremental knowledge transfer in object detection is made open-source and will be released for public distribution on Github. 
\end{itemize}

The rest of this paper is organized as follows. Section \ref{related} discusses the background on object detection methods and the relevant metrics for the evaluation of the detection accuracy. It also provides an overview of the related work. Section \ref{approach} explains the proposed system framework. In Section 4, the setup of experiments and the results are discussed. Finally, Section \ref{conclusion} concludes the paper.
\section{Background and Related Work}
\label{related}
In this section, the background on the object detection methods and the metrics for evaluating their accuracy are described. Different categories of related work are described and the drawbacks of each group are discussed.

\subsection{Background}

\noindent \textbf{Object detection methods:} Existing methods for object detection using CNNs can be classified as either two-stage or one-stage approaches. In two-stage methods such as FasterRCNN \cite{ren2015faster},  R-FCN \cite{dai2016r}, and AdaScale \cite{chin2019adascale} classification and localization are implemented using two separate steps involving classification and region proposal. The one-stage approaches (such as Yolo \cite{redmon2018yolov3}, SSD \cite{liu2016ssd}, and RetinaNet \cite{lin2017focal}) classify and localize objects in one step. One-stage detection models are generally faster while the accuracy of two-stage models is higher. However, at a smaller intersection of the ground-truth and the predicted object (intersection of union = 0.5), one-stage models can achieve nearly the same accuracy of the two-stage methods. In this paper we use single stage models, since they are better suited for embedded devices with limited computation resources.        

\noindent \textbf{Metrics for evaluation of detection accuracy:} There are three main validation metrics in object detection: Recall, Precision, and F1 score \cite{powers2011evaluation}. 

\noindent\textbf{Recall}: This is the number of correctly detected objects divided by the total number of objects in the scene. Recall is crucial in safety-critical systems where missing an object in the scene could be catastrophic.   

\noindent\textbf{Precision}: This is the total number of correctly detected objects divided by the total number of detected objects. This is useful for evaluating the systematic errors of detection. 

\noindent\textbf{F1 score}: This is a measure of overall detection accuracy and is defined as $2 \times \frac{Precision \times Recall}{Precision+Recall} $. 

\subsection{Related Work}

\subsubsection{Implementation of object detection on cloud} Neurosurgeon \cite{kang2017neurosurgeon} and JointDNN \cite{eshratifar2018jointdnn} are two recent examples of performing image classification collaboratively between an user-end device and a cloud server.  However, they do not consider object detection methods, which are a superset of image classification methods.  \textit{Glimpse} \cite{chen2015glimpse} performs object detection on mobile devices using a cloud server. When the network delay exceeds a certain threshold, their approach uses tracking to estimate the location of objects based on an active cache of frames. 

\textit{MobiEye} \cite{mao2019mobieye} is another cloud-based object detection system for mobile devices implemented in a multi-threaded asynchronous manner. The first thread sends the key frames to the cloud for object detection. The second thread performs the object tracking based on the result of processed key frames using optical flow network. The response time of object detection for key frames is dependent on the network delay. Therefore, when the network delay is high, the newly observed objects in the scene may be missed.    

\subsubsection{Domain adaptation} Domain adaptation refers to training a model for a specific domain of observation. There is a  substantial body of work on domain adaptation in object classification and detection methods \cite{lu2017unsupervised,chen2018domain,dai2018dark}. These methods have been presented to deal with challenges such as low quality images, and large variance in the background. This variation can result in a domain change in the training, validation, and test sets. However, existing methods do not consider sudden changes in the scene when performing inference at run-time. 
\subsubsection{Knowledge transfer}
Knowledge transfer (a.k.a knowledge distillation) \cite{mullapudi2019online,farhadi2019tkd} has been widely used to improve the accuracy of object detection systems. This method transfers the knowledge of a deeper model (oracle) to a lighter model (student) with fewer parameters. If the shallower model can gain this knowledge, it will have the same accuracy as the deeper model while using fewer resources. However, the student model would not adapt to this knowledge since it has fewer parameters but it can adapt partially.


Mullapudi et al. \cite{mullapudi2019online} proposed an image segmentation method where the shallow model is trained periodically at \ul{fixed} intervals. The selection of fixed interval may not be efficient. Farhadi et al. \cite{farhadi2019tkd} have proposed a systematic procedure using Long Short-Term Memory (LSTM) to determine the intervals at which the training of shallow model should be done. However, \ul{running the LSTM would be expensive on resource-confined user-end devices}.   

In this paper, a knowledge transfer framework is proposed in edge computing environment for energy-constrained user-end devices. This framework is explored in different ways using real-world direct measurements.

\section{System Framework}
\label{approach}
\begin{figure*}[]
    \centering
    \includegraphics[trim=0.0cm 3.0cm 0.1cm 2.9cm, clip=true,width=17.4cm]{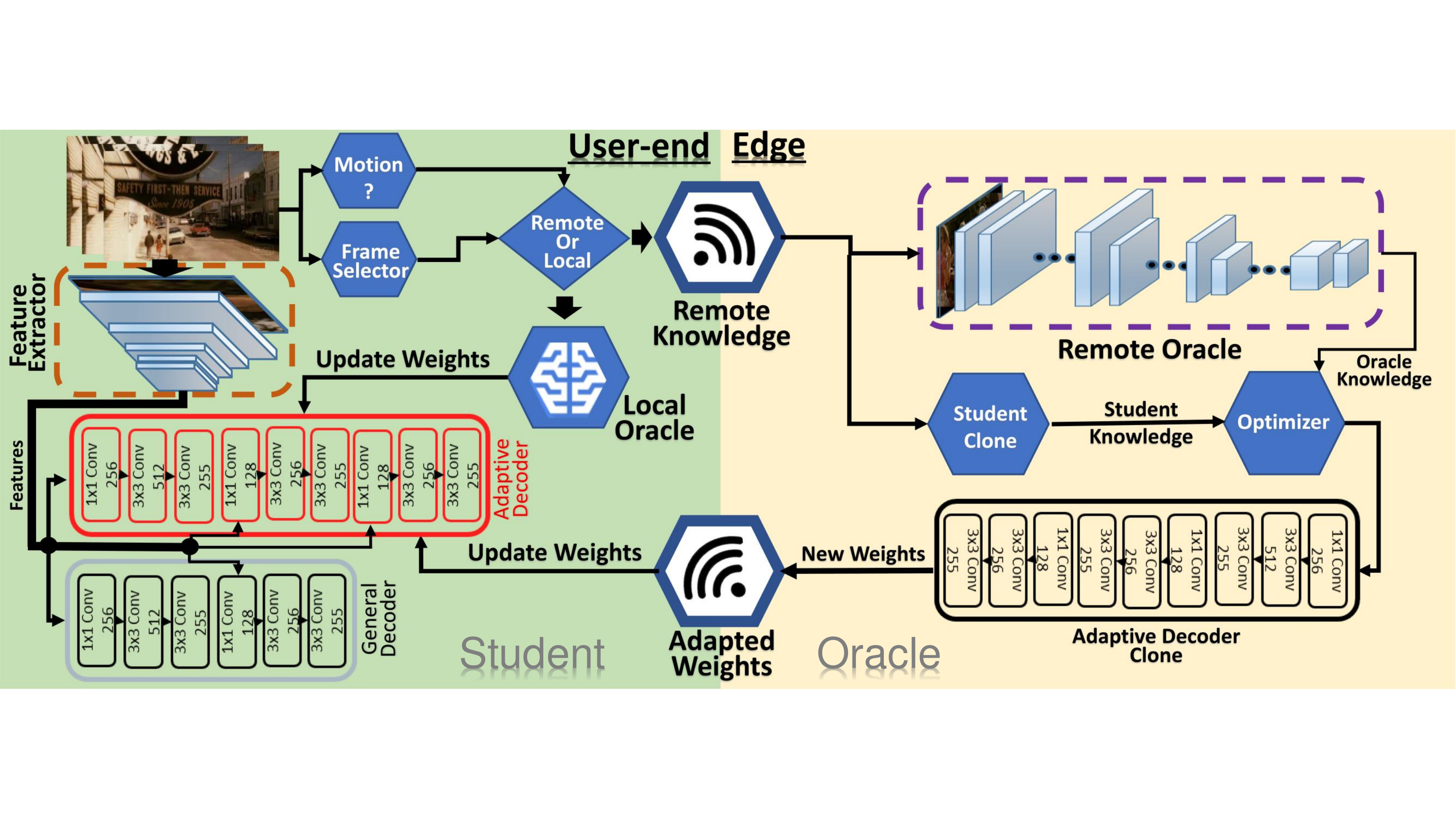}
    \caption{An overview of knowledge transfer method: The main thread of execution on the user-end device runs a shallow student model to detect objects. To keep the desired accuracy, a key frame selection module decides to retrain the student model based on the oracle model.}
    \label{TKD_fig}
\end{figure*}
\subsection{Knowledge Transfer}
In this section, the idea of online knowledge transfer in the edge computing domain is described. Moreover, an efficient method is proposed to select the frames that should be sent to the edge device for re-training the shallow model.  

In most vision datasets, there are a variety of domains that the model needs to learn. However, in real-world applications such as surveillance cameras, we are confronted with a specific domain and with limited types of objects. Although the scene can change,  the changes are typically slow. Here, the oracle knowledge over this temporal domain can be used to adapt the student model to the observing environment (which has been called temporal knowledge distillation \cite{mullapudi2019online}). This approach can improve the accuracy and response time. 

Although methods of knowledge transfer can improve the performance significantly, their implementation poses several challenges: 1) models may lose their generality and may not be able to detect objects seen for the first time; 2) training the student model for each and every incoming frame incurs high computation cost; 3) training a student model on the embedded device can affect other simultaneously running tasks, whereas training over the network incurs delay for adaptation. We address these challenges in the proposed approach.

\subsection{Main Architecture}

Our method of knowledge transfer between a shallow and deep model extends the approaches described in \cite{farhadi2019tkd,mullapudi2019online} to energy-constrained embedded systems. Figure \ref{TKD_fig} shows an overview of the proposed architecture which consists of three main parts: 1) the student model which is a shallow CNN model; 2) the oracle model which has a deep structure that can reach the state-of-the-art accuracy in detection; and 3) the key frame selection method which selects the epochs at which student adaptation occurs. 

The user-end device observes a scene and detects objects in the scene. Meanwhile, it adapts itself to the observing environment to improve the overall accuracy. In the following, each of these modules is described. Note that the presence of the oracle model in the user-end device is for evaluation purposes in our approach. 

\subsubsection{Student} 
This is a shallow model with a limited number of parameters requiring much less computation than a deep model. The student can learn only a limited amount of knowledge due to fewer parameters. In this paper, the student has a similar structure to Tiny-Yolo v3 \cite{redmon2018yolov3} with some modification to the decoder part. The student model (see Figure \ref{TKD_fig}) consists of three parts: 1) a feature extractor, 2) a general decoder, and 3) an adaptive decoder. The feature extractor, which has the same structure as the base model (Tiny-Yolo) is trained on a conventional object detection dataset. The general decoder (same structure as base model) detects objects using the extracted features and is trained along with a feature extractor during the training stage. Finally, the adaptive decoder is optimized during inference time using the oracle knowledge. Another stack of convolutions has been added to the adaptive decoder to improve the detection accuracy of small objects (Figure \ref{TKD_fig}).  
\subsubsection{Oracle} 

This is a model with a larger number of parameters and a deeper structure and as a result can reach the state-of-the-art accuracy over the target dataset (called oracle model). This model will be used to extract knowledge over the temporal domain. The knowledge will be used for adapting the student model to the observing environment. Here, Yolo v3 is used as the oracle model due to the similarity of structure to that of Tiny-Yolo and its lower latency compared with two-stage object detection methods. This makes the optimization procedure more efficient at inference time.

\subsubsection{Optimizer (weight update)} 
To transfer knowledge from the oracle to the student model, \textit{Adam} gradient descent method \cite{kingma2014adam} is used. First, the distance between the student and oracle model needs to be calculated. The student and oracle both have three output matrices with the same size. By calculating the $L2$ distance of student and oracle outputs ($\sum \lVert T_i^S - T_i^O \rVert_2^2 \ ,i=[1,2,3]$), the distance of two models is calculated. Next, using this distance as the loss value, the optimizer updates the weights of adaptive decoder in the student model. After adaptation, the weights of running student model will be replaced by the new weights.   

\subsubsection{Key frame selection} 
It is important to know when we need more knowledge while observing the environment. Training over all incoming frames will be expensive while training over a number of frames is needed to maintain accuracy. Mullapudi et al. \cite{mullapudi2019online} proposed a static interval key frame selection that is selected by the user at the beginning. To reach the desired accuracy, the interval should be small which will increase the inference time, energy consumption, and hardware utilization. Farhadi et al. \cite{farhadi2019tkd} proposed a combination of a uniform selector and an LSTM module to select key frames. This method achieves higher performance compared to Mullapudi et al. \cite{mullapudi2019online}. However, running the proposed LSTM module on a resource-constrained device would be costly. 

In our proposed approach, Kalman filtering \cite{kalman1960new} is used to track changes on the scene. If there is a significant change on the scene compared to the last adaptation, the frame is a good candidate for re-training. Also, frames are selected using a binomial selector described in equation \ref{eq:d}:  

\begin{equation}
\small
\setlength{\abovedisplayskip}{5pt}
\setlength{\belowdisplayskip}{5pt}
\begin{aligned}
& I \in  \{FALSE, \quad TRUE\} 
\\
& I = Motion(F_S ,F_L)\ \wedge \ I_R, \quad  I_R\sim B(2,P_t),
\\
& P_t=\left\{\begin{matrix}
 &max((P_{t-1}-0.05),0.05) \ \ \ \ \ \ \  \Delta L <\sigma,  &   \\ 
 &min(2P_{t-1},1.0) \ \ \ \ \ \ \ \ \ \ \ \ \ \ \ \ \ \ \ \ \  \Delta L >\sigma,   & 
\end{matrix}\right.
\\
 & F_S : Observed \ frame, F_L: Last \ key \ frame,
\end{aligned}
\label{eq:d}
\end{equation}

\noindent where $I$ denotes the final decision of the selector indicating whether or not to re-train. The $I_R$ is sampled from the binomial distribution. The probability of selection ($P_t$) is changed based on knowledge transfer loss. If $\Delta L$ (i.e. the difference between current loss and previous training loss), improves more than $\sigma$, the probability factor will be doubled to increase frames for knowledge transfer. Otherwise, $P_t$ will be decreased to select fewer frames for training. At least $5\%$ of frames are selected as a sample to avoid missing the scene changes. The value of $\sigma$ is a hyper parameter which is obtained after several rounds of experiments. This value can be changed based on the type of loss function. In this paper, the value of $\sigma$ is chosen to be $\sigma = 0.5$. 

Moreover, we will not select any other frame for adaptation while we are processing a key frame. This approach makes the detection system more adaptable to the newest observed objects by avoiding the queuing of frames.

\subsubsection{Knowledge source} 
The student module can be updated using local or remote knowledge. As shown in Figure \ref{TKD_fig}, the detection system has its own Oracle (local Oracle) to update the weights and adapt to the environment. The local Oracle is both optimizer and oracle model. Also, the user-end device can request a network oracle (edge) for a deeper knowledge and optimization. The edge has a clone of student model and updates this clone and sends it back to the user-end device. This approach is not on-time due to network latency. Section \ref{results} includes a detailed evaluation of these trade-offs.

\subsubsection{Multi-threading} 
To minimize the inter-effect of these sub-modules, each module is running on a separate thread. Although these modules are running independently, they may affect each other due to limited resources on the device. For instance, the oracle module can use most of the available CUDA cores while we are running the student model and cause an increase in inference time.

\section{Experimental Results}
\label{results}
In this section, the experimental setup and the results of applying our approach are described. 
\begin{figure}[b!]
    \centering
    \includegraphics[scale=0.075]{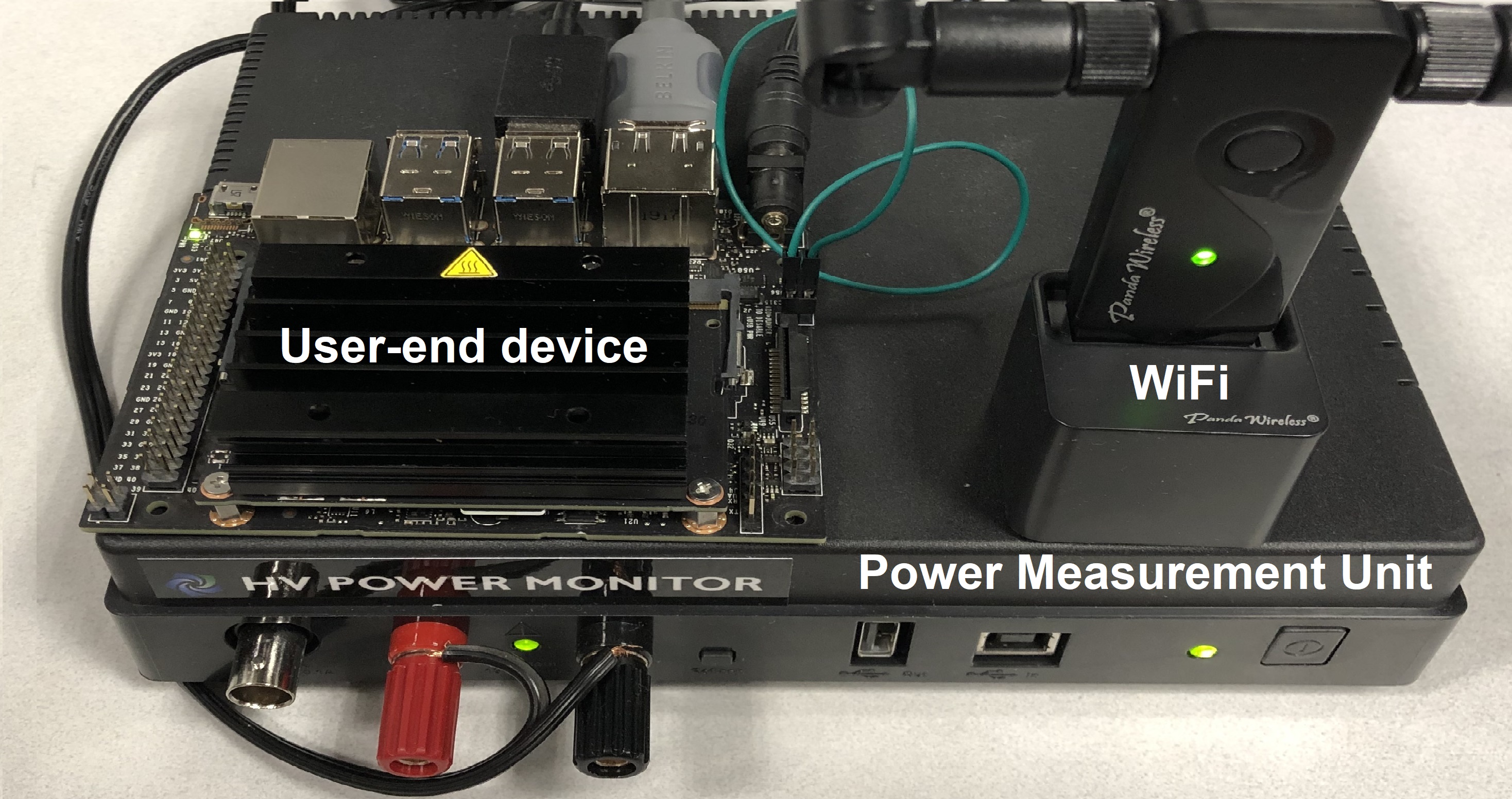}
    \caption{Experimental Setup.}
    \label{Setup}
\end{figure}

\subsection{Setup}
\noindent \textbf{Hardware Setup:} The user-end device is NVIDIA Jetson Nano developer kit \cite{jetsonnano} which is the first recipient of camera frames. This kit is equipped with a quad-core ARM A57 CPU operating at 1.43 GHz and a 4GB 64-bit LPDDR4 RAM. Moreover, it has a 128-core Maxwell GPU. The edge device is a Dell workstation with an Intel Xeon W-2125 CPU operating at 4.0 GHz having 32GB of RAM and an NVIDIA Titan Xp GPU. 

Both Local Area Connection (LAN) and Wi-Fi were tested as the communication medium between devices (tested in a network with a WAN (wide-area network) backbone). The bandwidth of Wi-Fi and LAN connections were approximately 13 and 100 Mb/s respectively. To measure the power consumption of the user-end device, Monsoon HV power monitor was used (Figure \ref{Setup}). 


\noindent \textbf{Dataset:} To verify the efficiency of the proposed approach, two types of videos from fixed and moving cameras were selected. The fixed camera case was from the surveillance videos in the UCF dataset \cite{sultani2018real}. The moving camera case was from the car crash dataset \cite{chan2016anticipating}. The initial weights of student and oracle model were obtained by training on the Microsoft COCO dataset \cite{lin2014microsoft}.     

\noindent \textbf{Evaluation Metrics:} The proposed approach to train on the edge device over a network was evaluated in terms of accuracy and performance metrics. The F1 and Recall scores are representative metrics for the accuracy of the detection method.

In our experiments, the output of deep model was assumed to be the ground-truth and the proposed approaches were compared with the knowledge of deep model. Moreover, the average inference time and training time for all the processed frames was measured. Additionally, the total energy consumption of the user-end device processing the whole video was compared. This energy consumption is attributed to different factors including video decoding, inference, non-maximum suppression (NMS) post processing, and network communication. In order to evaluate the overall efficiency of our approach, both in terms of accuracy and energy consumption, the overall score is used as the ratio of F1 score to energy consumption  \cite{DACcontest,alyamkin20182018}.

\begin{table}[]
\small
\centering
\label{table1}
\caption{Comparison of different approaches. Local training and network training both use the proposed key frame selection method using \ul{full precision data}. The energy column is the average energy consumption for each frame. Overall score is the ratio of F1 score to energy.}

\begin{tabular}{c|c|c|c|c|}
\cline{2-5}
                                                                                        & \multicolumn{4}{c|}{Metrics}                                                                                                                                                                                                       \\ \cline{2-5} 
                                                                                        & \begin{tabular}[c]{@{}c@{}}Energy\\ (J)\end{tabular} & \begin{tabular}[c]{@{}c@{}}Inference\\ Time (s)\end{tabular} & \begin{tabular}[c]{@{}c@{}}F1\\ Score\end{tabular} & \begin{tabular}[c]{@{}c@{}}Overall\\ Score\end{tabular} \\ \hline
\multicolumn{1}{|c|}{Shallow Model}                                                     & 1.06                                                 & \textbf{0.187}                                               & 0.489                                              & 0.46                                                    \\ \hline
\multicolumn{1}{|c|}{Deep Model}                                                        & 4.55                                                 & 0.669                                                        & 1                                                  & 0.22                                                    \\ \hline
\multicolumn{1}{|c|}{Local Training}                                                    & 1.83                                                 & 0.215                                                        & \textbf{0.753}                                     & 0.41                                                    \\ \hline
\multicolumn{1}{|c|}{\begin{tabular}[c]{@{}c@{}}Network\\ Training  (Wi-Fi)\end{tabular}} & 1.25                                        & 0.198                                               & 0.731                                              & 0.58                                           \\ \hline
\multicolumn{1}{|c|}{\begin{tabular}[c]{@{}c@{}}Network\\ Training  (LAN)\end{tabular}} & \textbf{0.98}                                        & \textbf{0.193}                                               & 0.745                                              & \textbf{0.76}                                           \\ \hline
\end{tabular}
\end{table}

\begin{figure*}[h]
    \centering
    \includegraphics[trim=5.0cm 8.2cm 6cm 4.4cm, clip=true,width=17.5cm]{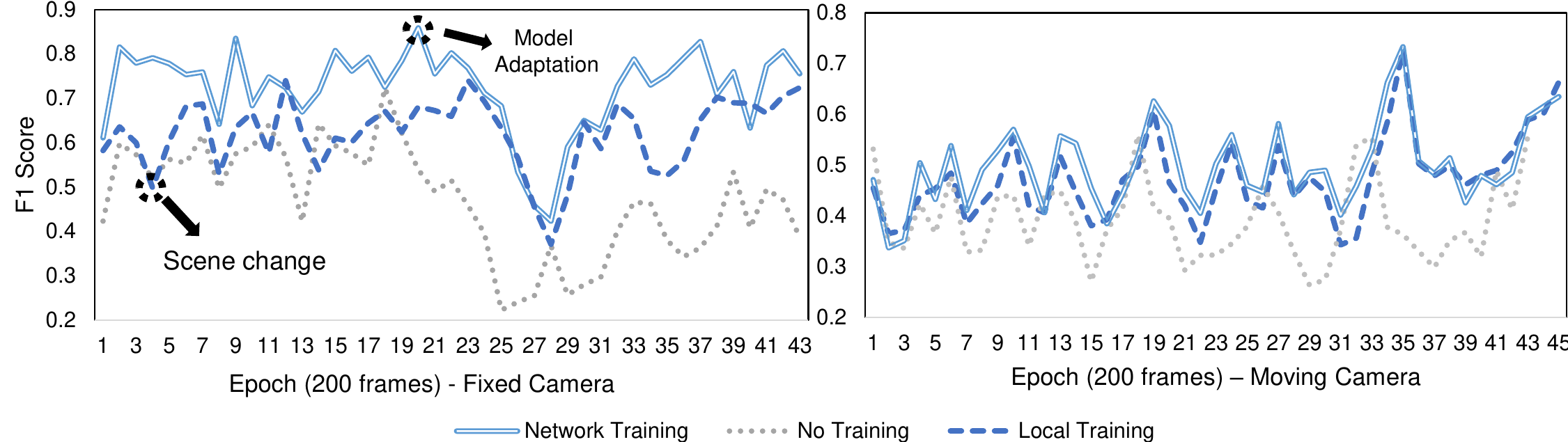}
    \caption{F1 score variation. In the case of fixed camera, the network training (NT) using Wi-Fi connection  even has a better F1 score in comparison with the local training (LT). Both NT and LT operate on \ul{half precision data}. High spike indicates that the model has been adapted to the environment while the low spike shows the scene change.}
    \label{variation}
\end{figure*}

\begin{figure*}[h]
    \centering
    \includegraphics[trim=0.5cm 2.0cm 7cm 2.0cm, clip=true,width=17.6cm]{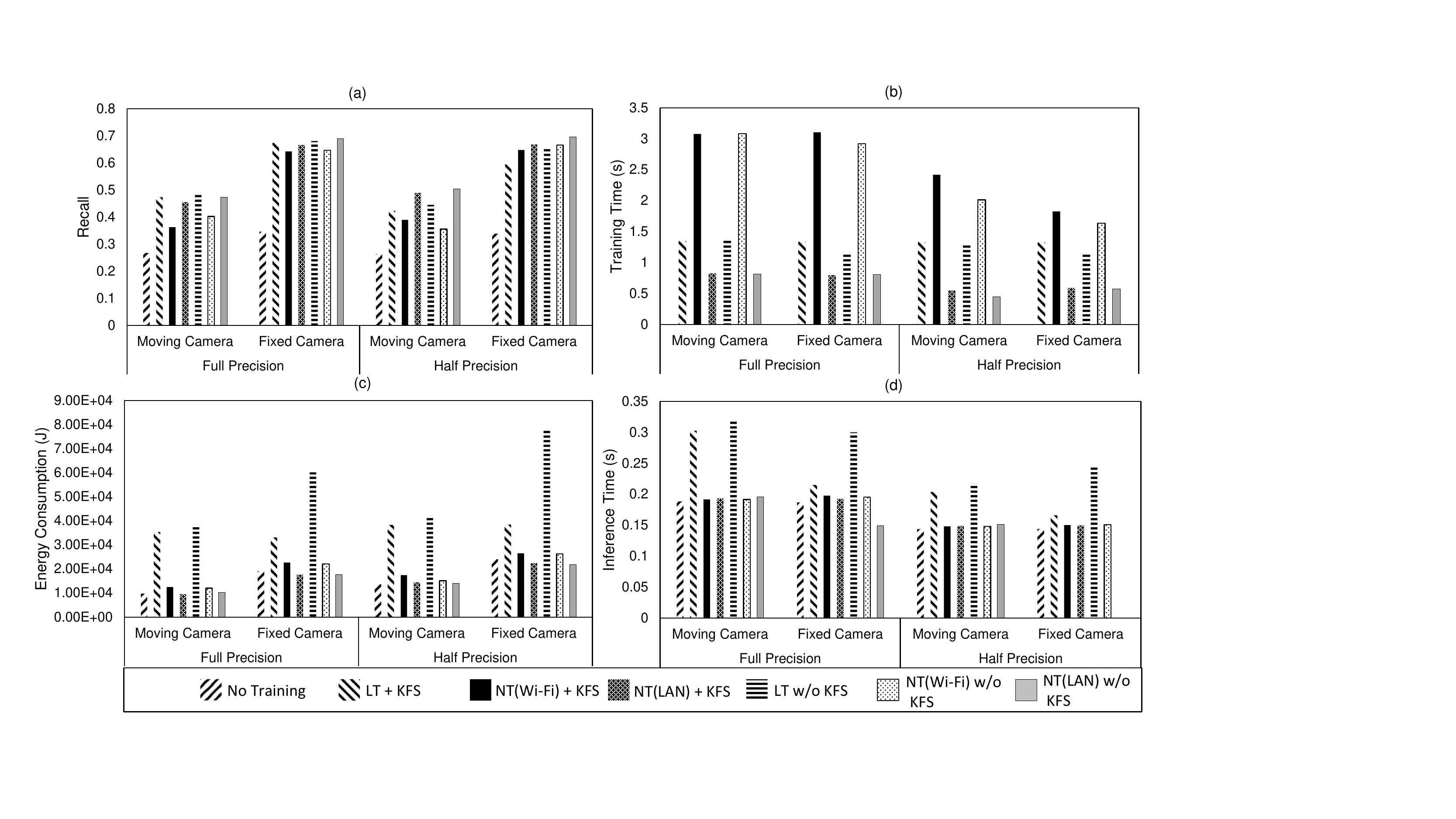}
    \caption{Comparison of (a) Recall, (b) Average training time, (c) Total energy consumption of all frames, (d) Average inference time, for fixed and moving camera videos using Wi-Fi and LAN connections. The efficiency of key frame selection method (KFS) has been also compared with the case in which all the frames are trained (w/o KFS).}
    \label{results}
\end{figure*}

\noindent \textbf{Test Scenarios:} The test scenarios include three cases: 
\begin{enumerate}
\item Network Training (\ul{NT}): the weights are updated on the edge device through either Wi-Fi or LAN connections. The input data is a tensor of size $416 \times 416\times 3$ which needs to be transmitted from the user-end device to the edge device. Moreover, the weights of decoder (explained in Figure \ref{TKD_fig}) will be sent from the edge to the user-end device. 
\item Local Training (\ul{LT}): The weights are updated locally on the user-end device. In this approach, the calculation of loss function to update the weights is also done on the user-end device.
\item No Training: The weights are not updated at all and the inference is done using the shallow model.  
\end{enumerate}
The aforementioned scenarios were tested using full precision and half precision weight and data values on the videos of both fixed and moving cameras. The efficiency of key frame selection (\ul{KFS}) method was compared with the case in which all the frames were selected as the key frame (\ul{w/o KFS}).
\subsection{Experimental Results}
 Table I shows the comparison of LT and NT versus running the deep or shallow model on the user-end device for the fixed camera video. In case of NT using LAN, the energy consumption and inference time were reduced by around 78\% and 71\% compared with running the whole deep model on the user-end device while the F1 score was reduced only by $ \approx 25\%$. However, as shown in \cite{farhadi2019tkd}, the accuracy of student model using knowledge transfer remains almost the same as the deep model in Youtube video dataset \cite{prest2012learning}. On the other hand, running the shallow model on the user-end device resulted in unacceptable accuracy (F1 score = 0.489). Note that training the shallow model locally led to additional 87\%  energy consumption. The overall efficiency of the approaches was evaluated using the overall score (i.e. the ratio of F1 score to energy). The overall score of network training (LAN) leads to 1.65x and 3.45x improvement compared with running the shallow and deep model on the user-end device. The reason for getting better overall score in LAN compared with Wi-Fi is due to the lower communication delays in the network. Still, the network training using Wi-Fi can gain better overall score compared with other approaches.

The variation of F1 score was also measured (shown in Figure \ref{variation}). Both NT and LT perform significantly better than the case in which no incremental training was done (both fixed and moving camera videos).  Moreover, higher F1 score was achieved in the case of fixed camera videos. It demonstrates that the adaptation is more effective in fixed camera videos due to fewer changes in the scene. In the case of fixed camera, NT outperforms LT since training latency is lower which leads to higher achievable accuracy.

The presence of low spikes in some parts of Figure \ref{variation} suggests that there is are significant scene changes in the video at those epochs. On the other hand, the high spike indicates the epochs that the detection system was able to adapt to the environment. Due to different re-training latency values, a shift in the spikes among different scenarios can be observed.

Figure \ref{results} (a) shows the average Recall for the different scenarios mentioned earlier. Recall in the case of LAN connection was better than Wi-Fi. This is due to the fact that the data transmission latency of LAN connection was lower than Wi-Fi and the student model was updated more frequently.  On the other hand, the accuracy of NT and LT are close even when using the low speed wireless connection since the training latency is the same. Moreover, training over the network using half precision data reduces the communication time for sending the data and weight values significantly while having a negligible effect on the accuracy. Note that the values of F1 score followed the same trend as with Recall.

Figure \ref{results} (b) shows the average training time of all processed frames. Network training using a LAN connection has the lowest training time while  training using the Wi-Fi has the highest among all cases. Using half precision data reduces the training time significantly specifically in NT using Wi-Fi connection. Even using half precision data, LT was still having lower training time than the NT (Wi-Fi) since the training time locally using half precision incurs lower computation.  

The total energy consumption of the user-end device for different scenarios is compared in Figure \ref{results} (c). The lowest energy consumption was achieved using NT with LAN connection. This is due to the fact that the transfer of weights to the edge device was done faster and the user-end device was not involved in the training procedure. Although the training time for NT (Wi-Fi) was highest, the energy consumption is close to NT(LAN) case (NT (LAN) achieves higher accuracy as mentioned before). It can be also seen that LT leads to higher energy consumption compared with NT. There are two reasons for this observation: 1) The high computation cost of running the oracle model locally; 2) Increase of inference time due to interference of local training with the online inference.

Figure \ref{results} (d) shows the average inference time of all frames. The effect of local training can be again observed in the higher inference time compared with the network training. NT(Wi-Fi) leads to lower inference time compared with NT(LAN). The reason is due to higher accuracy obtained in NT(LAN) scenario. Note that the post processing on the detected objects (NMS) took longer time for the scenarios with higher accuracy which led to higher inference time.  

The efficiency of key frame selection method was also evaluated in the experiments. It can be observed in Figure \ref{results} (a) that NT (LAN)+KFS performs closely in terms of Recall metric compared with the case where the training happens at all video frames (NT (LAN) w/o KFS). Moreover, the energy consumption and the inference time in the case of LT w/o KFS is significantly higher in comparison with LT+KFS since the re-training should happen for all the frames in LT w/o KFS.   
\subsection{Further Discussion}
The experimental results gave us some insights on how to design the system for the implementation of online knowledge transfer. The takeaways can be summarized as below:
\begin{itemize}
    \item Local training: The energy consumption of this method is higher compared to other approaches. On the other hand, the training time is more predictable in comparison with network training using Wi-Fi. This is due to the fact that the communication time using Wi-Fi follows a stochastic behavior. 
    \item Network training: This approach can lead to higher accuracy/energy ratio on average. Network training using LAN connection is as predictable as local training since the LAN connection is more stable compared with Wi-Fi. 
    \item Loss Function: The used loss function in this paper is based on the Euclidean distance of student and oracle model. However, the calculation of this loss function is computationally intensive on the CPU of user-end device. Therefore, local training can be more expensive in terms of energy consumption and inference time. 
    \item Frame selection: The frame selection strategy selects more frames to train at the beginning of the environment observation. The number of selected frames is decreased throughout the observation of the environment. This observation was more vivid in fixed camera videos. 
    \item Potential solution: Based on these observations, a combination of network and local training is suggested where the initial frame training can be done on the edge device. For the rest of the frames, a decision making policy can determine whether the training should be done locally or on the edge device. The decision making policy can be based on the stability of the communication media. Moreover, the number of frames to be trained affects the decision making policy. For instance, when the number of frames to be trained is higher, the network training is a better option.    
\end{itemize}





\section{Conclusion}
\label{conclusion}
In conclusion, we designed and implemented a framework for incremental knowledge transfer in edge computing environment. The parameters of a shallow model running on the user-end device are updated during inference at some key frames to achieve the close accuracy as using a deep model. We demonstrated the proposed approach in the real-world scenario. Our framework consisting of a shallow and a deep model resulted in 78\% energy reduction when compared to running the deep model alone. The experiments also revealed that the latency of communication must be accounted for when deciding to do the model updates.

\noindent\textbf{Acknowledgment:} The National Science Foundation under the Robust Intelligence Program (\#1750082) and I/UCRC Center for Embedded Systems (\#1361926), the IoT Innovation (I-square) fund provided by ASU Fulton Schools of Engineering are gratefully acknowledged.

\bibliographystyle{IEEEtran}
\bibliography{References}

\end{document}